\documentclass[12pt]{l4dc2021} 

\usepackage{graphicx}      %
\usepackage{siunitx}

\usepackage{soul}
\usepackage{color}
\usepackage{floatrow}
\usepackage{stfloats} 
\usepackage{dsfont}

\usepackage{float}
\floatstyle{plaintop}
\restylefloat{table}

\title[Nonlinear state-space identification using deep encoder networks]{Nonlinear state-space identification using deep encoder networks}
\usepackage{times}

\author{\Name{Gerben Beintema} \Email{g.i.beintema@tue.nl}\\
 \Name{Roland Toth} \Email{r.toth@tue.nl}\\
 \Name{Maarten Schoukens} \Email{m.schoukens@tue.nl}\\
 \addr Department of Electrical Engineering, Eindhoven University of Technology, 5600 MB, The Netherlands
}

\begin{document}

\maketitle

\begin{abstract}
Nonlinear state-space identification for dynamical systems is most often performed by minimizing the simulation error to reduce the effect of model errors. This optimization problem becomes computationally expensive for large datasets. Moreover, the problem is also strongly non-convex, often leading to sub-optimal parameter estimates. 
This paper introduces a method that approximates the simulation loss by splitting the data set into multiple independent sections similar to the multiple shooting method. 
This splitting operation allows for the use of stochastic gradient optimization methods which scale well with data set size and has a smoothing effect on the non-convex cost function.
The main contribution of this paper is the introduction of an encoder function to estimate the initial state at the start of each section. The encoder function estimates the initial states using a feed-forward neural network starting from historical input and output samples. The efficiency and performance of the proposed state-space encoder method is illustrated on two well-known benchmarks where, for instance, the method achieves the lowest known simulation error on the Wiener--Hammerstein benchmark.
\end{abstract}


\section{Introduction}

Linear system identification is already well developed, in both a theoretical and a practical sense. However, due to increasing performance demands, the use of light-weight materials and/or increased demand for energy efficiency, linear identification and control falls short of meeting these demands and hence, nonlinear system identification and control has become increasingly important. This paper introduces a novel nonlinear state-space identification approach that reduces the computational cost by combining techniques and insights of machine learning and dynamical system identification. 

Computational tractability is often hard to achieve while estimating nonlinear (state-space) models when minimizing the simulation error. Nevertheless, the use of simulation error is essential in practise to increase model reliability when model errors are present~\citep{schoukens2019nonlinear}. However, the simulation error objective is commonly computationally intractable often caused by the lack of smoothness of the loss function and/or the gradients of the loss function~\citep{ribeiro2019smoothness}. Furthermore, the computational cost of calculating the loss scales linearly with the length of the measured time-series. This limits the applications to relatively small datasets and is a serious detriment in the age of big data. Moreover, it has been shown that artificial neural network are powerful function approximators when large datasets are applied~\citep{chiroma2018ANNprogress}.

Multiple methods exist which aim to negate these causes of intractability: (\textit{i}) Careful initialization of the model parameters~\citep{NLLFR}, resulting in an initial estimate which is expected to be close to the global minimum of the loss function and avoids gradient and system instability, and (\textit{ii}) the multiple shooting method~\citep{multi-shoot-old} which splits the time series into multiple sections where each section has its own independent loss function. This splitting operation has recently been shown to have a smoothing effect on the loss function and its gradient~\citep{ribeiro2019smoothness} making gradient based optimization method easier to apply.

How to estimate the initial state at the start of each section remains one of the main issues in successfully applying the multiple shooting method. Two approaches are commonly used: (\textit{i}) Setup the initial states as parameters of the optimization~\citep{multi-shoot-old}, this however scales the model complexity with the number of sections, and (\textit{ii}) estimate the initial state by using equality constrains to the final state of the previous section~\citep{ribeiro2019smoothness}, this constraint optimization is considerably more involved.

This paper proposes a new approach to the initialization problem by using an encoder function. This function estimates the current state based on historical inputs and outputs. The use of an encoder function in combination with a multi-step ahead prediction loss can be viewed as an extension of sub-space identification approaches such as CCA~\citep{Katayama2007subspacebook}. The proposed approach reduces the transient errors and does not increase the model complexity with the number of sections and, moreover, it provides generalization to other unseen datasets as the encoder allows one to jump start the simulation at the correct model state. Furthermore, in this paper we demonstrate that the state-space encoder method with artificial neural network 
achieves state-of-the-art performance on the Wiener--Hammerstein benchmark~\citep{schoukens2009wiener} and obtains a competitive performance on the Silverbox benchmark~\citep{benchmark} significantly exceeding previously proposed deep learning methods on these benchmarks.\footnote{Code available at \url{https://github.com/GerbenBeintema/SS-encoder-WH-Silver}}

The remainder of this paper first discusses and motivates the state-space encoder method in Section \ref{sec:encoder}. Next, the method is applied to two well-known benchmarks and the results are compared quantitatively with other results from literature in Section \ref{sec:num} followed by a discussion of the presented approach in Section \ref{sec:dis}. \vspace{-0.25cm}

\section{Encoder-Based nonlinear state-space identification}
\label{sec:encoder}

\subsection{State-space model structure}

The following discrete-time model structure is considered: \vspace{-0.25cm}
\begin{subequations}
\begin{align}
    \hat{x}_{t+1} &= f_\theta(\hat{x}_t,u_t), \\
    \hat{y}_t &= h_\theta(\hat{x}_t,u_t),
\end{align}
\end{subequations}
where $t \in \mathds{Z}$ is the discrete time index, $x\in \mathds{R}^{n_x}$ the internal state vector, $u_t \in \mathds{R}^{n_u}$ and $y_t \in \mathds{R}^{n_y}$ are the model input and output respectively, $\theta$ the model parameters, and $f_\theta$, $h_\theta$ are the nonlinear dynamics of the state-space model. We assume that the measured data is generated by a system contained within this model class: $y_t = h_{\theta_0}(x_t,u_t) + v_t$ and $x_{t+1} =  f_{\theta_0}(x_t,u_t)$, where $v_t \in \mathds{R}^{n_y}$ is zero-mean (possibly coloured) noise with finite variance.

\subsection{Classical simulation error identification}

A common approach to estimate the model parameters is to minimize the simulation error as \vspace{-0.25cm}
\begin{equation}
    V_{\text{simulation}}(\theta) = \frac{1}{N_{\text{samples}}} \sum_{t=1}^{N_{\text{samples}}} ||h_\theta(x_t,u_t) - y_t||_2^2.
    \label{eq:OE}
\end{equation}
where $x_t$ dependent on $f_\theta$. For convenience of writing use a single parameter vector $\theta$ for both functions, however, in practise they are independently parameterized. However, using this expression directly is often challenging. Each evaluation of $V_{\text{simulation}}(\theta)$ requires $O(N_{\text{samples}})$ operations in series which makes it intractable for large data sets. Moreover, it is known empirically and shown theoretically that this expression can result in many local minima and in unstable behaviour of the optimization for gradient based techniques~\citep{ribeiro2019smoothness}. 

\subsection{Encoder networks for nonlinear identification}

\begin{figure}
    \centering
    \includegraphics[width=0.6\linewidth]{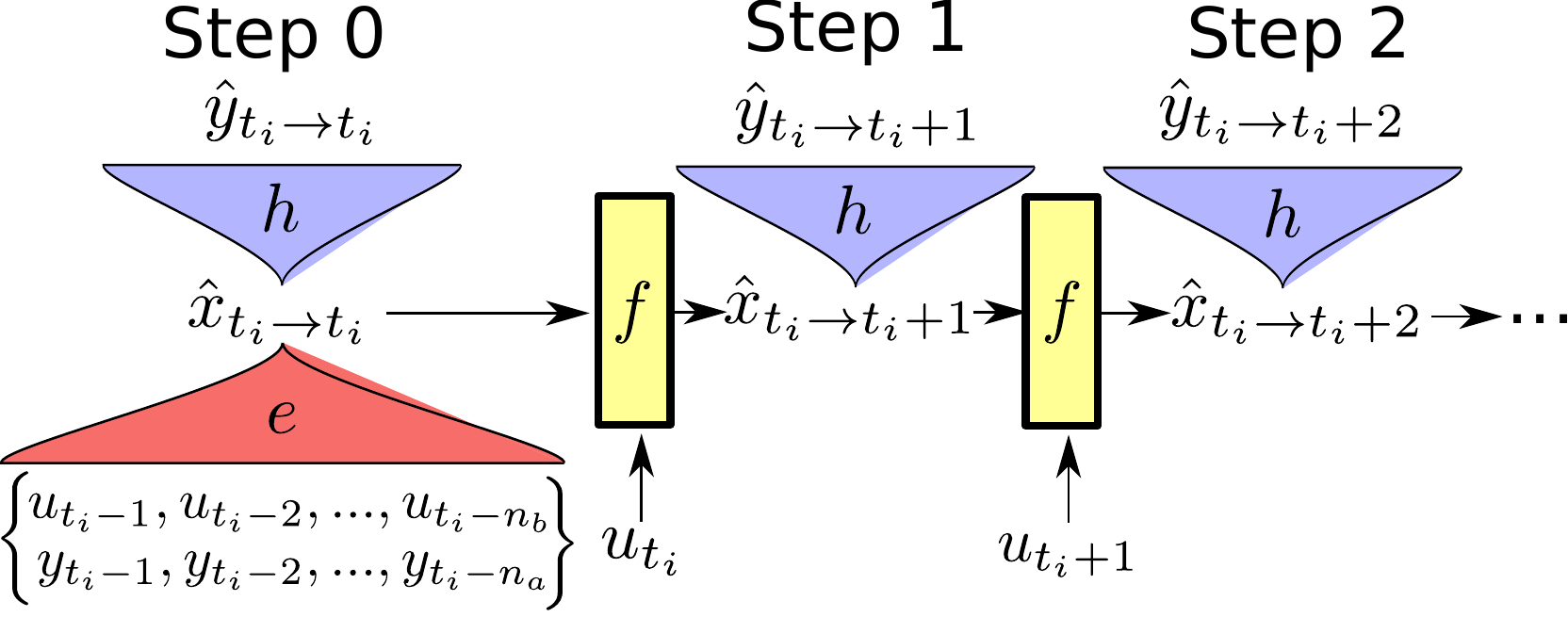}
    \vspace{-4 mm}
    \caption{The state-space encoder method applied on a section of a time series starting at $t_i$ where the initial state $\hat{x}_{t_i \xrightarrow{} t_i}$ is estimated using the encoder function $e$ using the historical inputs and outputs.}
    \label{fig:encoder}
\end{figure}

To negate the problems observed for the simulation loss we employ a loss function that sums over $N$ independent sections of the data with starting index $t_i$ and length $T+k_0+1$ similar to the multiple shooting method which is known to have a stabilizing and smoothing effect~\citep{ribeiro2019smoothness}. The full formulation of the proposed state-space encoder method is given by:\vspace{-0.25cm}
\begin{subequations}
\begin{align}
    V_{\text{encoder}}(\theta) &= \frac{1}{2 N (T+1)} \sum_{i=1}^N \sum_{k=k_0}^{T+k_0} ||\hat{y}_{t_i \xrightarrow{} t_i + k} - y_{t_i+k}||_2^2, \\
     \hat{y}_{t_i \xrightarrow{} t_i + k} &:= h_\theta( \hat{x}_{t_i \xrightarrow{} t_i +k}, u_{t_i + k}),\\
     \hat{x}_{t_i \xrightarrow{} t_i + k+1} &:= f_\theta(\hat{x}_{t_i \xrightarrow{} t_i + k},u_{t_i + k}),\\
     \hat{x}_{t_i \xrightarrow{} t_i} &:= e_\theta(y_{t_i-n_a:t_i-1},u_{t_i-n_b:t_i-1}),
     \label{eq:encoder}
\end{align}
\end{subequations}
where $t_i \xrightarrow{}  t_i + k$ reads as ``The simulated state at $t_i + k$  starting at $t_i$ with initial state $\hat{x}_{t_i \xrightarrow{} t_i}$''. Furthermore, the choice of $k_0 \geq 0$ allows for an initial transient time to be excluded from the loss calculation. Finally, to close this expression, an encoder function $e_\theta$ is introduced to estimate the initial state starting from historical input and output samples $u_{t-n_b:t-1} \in \mathds{R}^{n_u \cdot n_b}$ and $y_{t-n_a:t-1} \in \mathds{R}^{n_u \cdot n_a}$. A graphic representation of the state-space encoder method is shown in Figure \ref{fig:encoder}. 

This approach and multiple shooting is related to Truncated Backpropagation Through Time (TBTT)~\citep{tallec2017TBTT} which is a gradient computation method which truncates the gradient calculations after a few backwards steps. This, however, requires a initial pass over the entire dataset length which similarly to simulation error loss (Equation \eqref{eq:OE}) scales the computational complexity with dataset length. Moreover, Our approach and multiple shooting works on the level of the cost function whereas TBTT is a gradient computation method. 

This expression can be interpreted as a trade-off between simulation error with $N=1$, $T=N_{samples}$ and prediction error with $T=0$, $N=N_{samples}$, $k_0 = 0$ under the right assumptions. Note that sections can overlap. This is normally excluded in multiple shooting approaches, but is allowed in the approach presented in this paper. Neural network structures (e.g. fully connected neural networks, convolutional neural networks) are used to represent the encoder function due to their excellent function approximation abilities for large data sets~\citep{nn-approx}.

This approach has a few computational advantages over the two existing state initialization methods when considering the multiple shooting method context. As mentioned in the introduction, a first method is to initialize as $\hat{x}_{t_i \xrightarrow{} t_i} = 0$~\citep{multi-shoot-old} which often requires a system-specific burn time $k_0$ such that the transient is sufficiently suppressed. This burn time can significantly increase the computational cost when a long or even infinite transient (e.g. resonating and chaotic systems) is present. The second method includes $\hat{x}_{t_i \xrightarrow{} t_i}$ as a model parameter~\citep{ribeiro2019smoothness}. However, this significantly increases the model complexity by adding $N \cdot n_x$ parameters. Moreover, this method also provides no generalization to other datasets. The proposed encoder method negates the above mentioned computational hindrances and potentially generalizes to new data sets.

The state-space encoder method also connects with the sub-space identification literature. Sub-space identification uses an encoder and a decoder map to identify a sub-space for a dynamical system such as CCA~\citep{Katayama2007subspacebook}. This approach can be extended to the proposed method by choosing the structure of the decoder map to an unrolled state-space model as in Figure \ref{fig:encoder}. This connection will be further explored in future research. Moreover, the proposed method is also arguably simpler than the closely related auto-encoder approach~\citep{app:auto-masti2018learning} for it only employs a single loss function and it allows for the minimization of a multi-step criterion. \vspace{-0.33cm}

\subsection{Batch optimization}

Adapting the cost framework proposed in Equation \eqref{eq:encoder} allows the loss to be calculated independently on each section. Firstly, this independence allows for close to trivial parallelization, resulting in a reduced computational cost on modern hardware. Secondly, one can choose to sum not over all sections but only a subset of possible sections. This results in a batch loss formulation of the multiple shooting method as:\vspace{-0.4cm}
\begin{subequations}
\begin{align}
    V_{\text{batch}}(\theta) &= \frac{1}{2 N_{\text{batch}} (T+1)} \sum_{i \in B } \sum_{k=k_0}^{T+k_0} ||\hat{y}_{t_i \xrightarrow{} t_i + k} - y_{t_i+k}||^2, \\
    B &\subset \{1,2,...,N\}.
\end{align}
\end{subequations}
The batch loss formulation allows one to utilize modern powerful batch optimization algorithms developed by the machine learning community (e.g. Adam \citep{adam}) that scales well for increasing data set size. \vspace{-0.4cm}

\section{Numerical experiments}
\label{sec:num}
In this section, the real-world modeling performance of the state-space encoder method is analyzed by applying the proposed method to two well-known system identification benchmarks: The Wiener--Hammerstein and the Silverbox benchmark.\footnote{Data obtained from \url{https://sites.google.com/view/nonlinear-benchmark/}} The obtained results are compared quantitatively with other results listed in the literature. 

\subsection{The Wiener--Hammerstein benchmark}
\label{sec:wien}

The Wiener--Hammerstein benchmark~\citep{schoukens2009wiener} is implemented as an electronic circuit with a diode-resistor nonlinearity (SISO). This benchmark consists of 80,000 training samples, 20,000 validation samples and 78000 test samples.

The encoder function $e_\theta$ and both dynamics function $f_\theta$ and $h_\theta$ are all represented using a single hidden layer neural network with 15 hidden nodes and tanh activation functions. Furthermore, this structure also includes a parallel linear function that goes directly from the input of the neural network to the output of the network without any nonlinear activation functions similar to a residual layer~\citep{residual}: \vspace{-0.44cm}
\begin{equation}
    \mathbf{z}_{\text{out}} = \mathbf{A}_1 \tanh(\mathbf{A}_2 \mathbf{z}_\text{in} + \mathbf{b}_2) + \mathbf{A}_3 \mathbf{z}_\text{in} + \mathbf{b}_3
\end{equation}
with $\mathbf{z}_\text{in}$ being the network input in vector form, $\mathbf{z}_\text{out}$ the network output and $\mathbf{A}_i$ and $\mathbf{b}_i$ the parameters of the network. These parameters are initialized by sampling from the uniform distribution $\mathcal{U}(-\sqrt{k},\sqrt{k})$ with $k=1/\sqrt{n_{\text{in}}}$ where $n_{\text{in}}$ are the number of inputs (i.e. number of elements in $\mathbf{z}_\text{in}$)

The Wiener--Hammerstein benchmark encoder state-space model structure has the following settings $n_x = 6$ (equal to the underlying system order), $k_0 = 0$ (no transient corrections required), $T = 80$ (taken approximately four times the time scale of the system which is approximately 20 steps), $n_a = n_b = 50$ (larger than $n_x$ and a few time the time scale). Furthermore, a 32-bit floating-point accuracy is used for the parameter estimation. The multiple shooting starting points $t_i$ can be any possible starting point within the range of the training set to maximally use the available data. This allows for overlapping training sections. Furthermore, the Adam batch optimization method~\citep{adam} is utilized with a learning rate of $\alpha = 10^{-3}$ and a batch size of $1024$ which adjusts the learning rate based on the variance of the gradient. During optimization the performance of the estimated model is evaluated by monitoring the simulation error on the validation set after each epoch. The estimated model is saved if a new lowest simulation error on the validation set has been achieved (i.e. early stopping). Furthermore, both the input and the output are normalized by subtracting the mean and dividing by standard deviation to improve the performance and training time. Lastly, after the batch training converged, a local minimum search using all training data is performed. However, this has only improved the final result marginally (i.e. $0.101\%$ test NRMS without final search).

The model performance is reported in both Root Mean Square (RMS) and the Normalized Root Mean Square (NRMS) of the simulation error:
\begin{equation}
    \text{NRMS} = \frac{\sqrt{1/N \sum_{t=t_0}^{N+t_0} ||\hat{y}_t - y_t||_2^2} }{\sigma_y} = \frac{\text{RMS}}{\sigma_y}
\end{equation}
with $\sigma_y = 244.7 \ \si{mV}$ the standard deviation of the measured test output.

\newcommand{\doubleline}[1]{\begin{tabular}[c]{@{}l@{}}#1\end{tabular}}
\newcommand{\tripleline}[1]{\begin{tabular}[c]{@{}l@{}l@{}}#1\end{tabular}}

\begin{table}[t]
\centering
\caption{Performance of the state-space encoder on the Wiener--Hammerstein benchmark compared to the results reported in the literature.}
\vspace{-5 mm}
{\small 
\begin{tabular}{l|ll}
Identification Method               & \doubleline{Test RMS\\ Simulation ($\si{mV}$)} & \begin{tabular}[c]{@{}l@{}}Test NRMS\\ Simulation\end{tabular} \\ \hline
\textbf{State-space Encoder} (this work)       & \textbf{0.241}                                                     & \textbf{0.0987\%}                                               \\
QBLA \citep{app:QBLA-schoukens2014identification} & 0.279                                                           & 0.113\%                                                        \\
Pole-zero splitting \citep{app:pole-splitting-sjoberg2012identification}  & 0.30                                                               & 0.123\%                                                        \\
NL-LFR \citep{NLLFR} & 0.30                                                               & 0.123\%                                                        \\
PNLSS \citep{app:PNLSS-marconato2009identification} & 0.42                                                               & 0.172\%                                                        \\
{Generalized WH \citep{app:gen-WH-wills}}      & 0.49                                                               & 0.200\%                                                        \\
LS-SVM \citep{app:ls-svm-falck2009identification}     & 4.07                                                               & 1.663\%                                                        \\
{Bio-social evolution  \citep{app:bio-naitali2016wiener}} & 8.55                                                               & 3.494\%                                                        \\
{Auto-encoder (reproduction) \citep{app:auto-masti2018learning}} & 12.01                                                              & 4.907\%                                                        \\
{Genetic Programming \citep{app:GP-Dhruv}} & 23.50                                                              & 9.605\%                                                        \\
{SVM \citep{app:svm-marconato2009identification}}                 & 47.40                                                              & 19.373\%                                                       \\
BLA \citep{app:lin-lauwers2009modelling}                  & 56.20                                                              & 22.969\%                                                      
\end{tabular}
}
\label{tab:result}
\end{table}

The results obtained on the Wiener--Hammerstein benchmark are reported in Table~\ref{tab:result}. The table shows that the proposed encoder method has, to the author's knowledge, the best known RMS simulation error reported in the literature for this benchmark. Furthermore, one can see in Figure~\ref{fig:time-feq} that the remaining error in the time domain is visually a straight line and the remaining error in the frequency domain is reduced significantly compared to error obtained by the best linear approximation. Also note that other models with larger neural networks and higher state dimension $n_x$ resulted in a similar model performance (e.g. with $n_x=8$ and two hidden layer neural networks yielded a test NRMS simulation of $0.1011\%$). This insensitivity to the model structure setting indicates that the importance of careful model structure selection is reduced for the state-space encoder method when being used in combination with large data sets. Note that the generalization gap (i.e. the gap between training error and test error) is negligible. The NRMS simulation error on the training set is $0.09789\%$ and on the test set is $0.09870\%$. This indicates that virtually no overfitting is taking place.
Also observe that these results are obtained using random initial parameters while many of the approaches listed in Table~\ref{tab:result} require a linear model estimate or other parameter initialization schemes to obtain competitive results.

\begin{figure}[tbp]
\floatconts
{fig:time-feq}%
{\vspace{-10 mm} \caption{The simulation error of the state-space encoder method evaluated on the test set of the Wiener--Hammerstein benchmark in both time and frequency domain.}}%
{%
\subfigure[Time domain][t]{%
\label{fig:pic1}%
\includegraphics[width=0.46\textwidth]{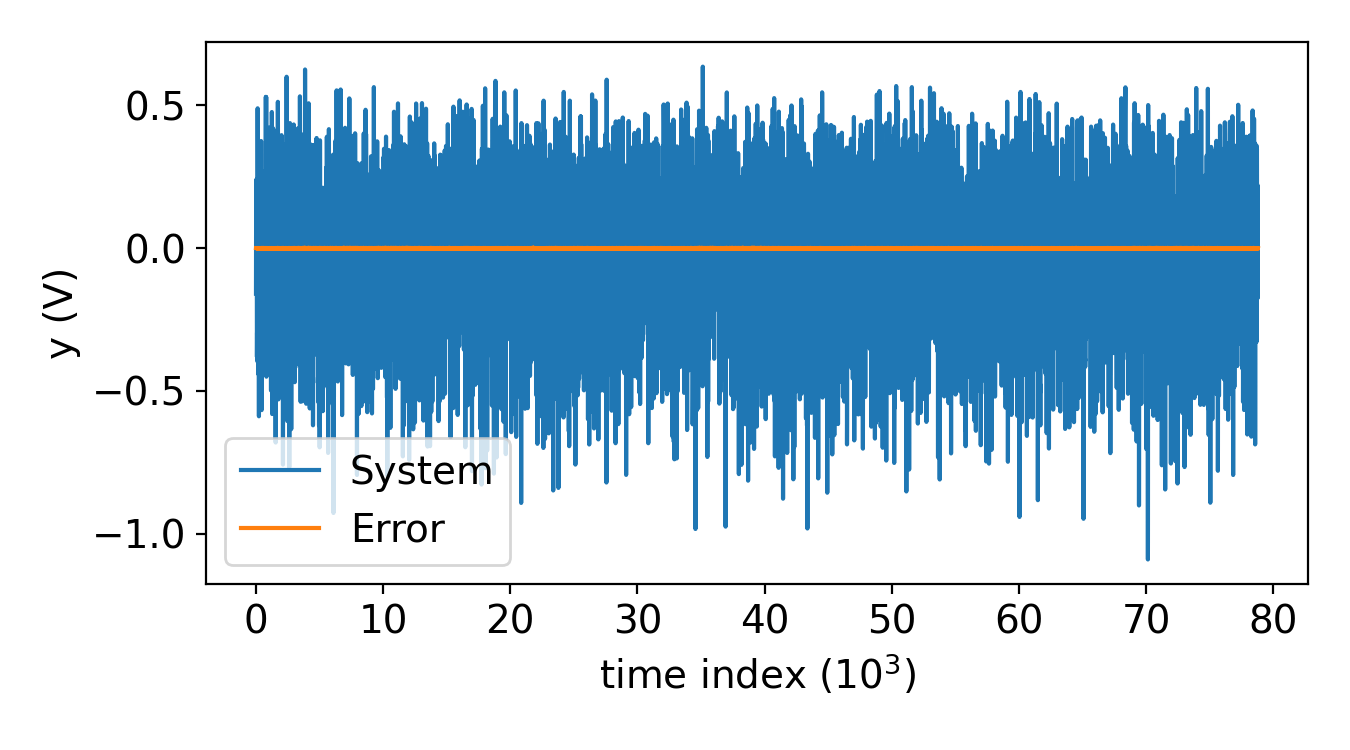}
}\qquad %
\subfigure[Frequency domain][t]{%
\label{fig:pic2}%
\includegraphics[width=0.46\textwidth]{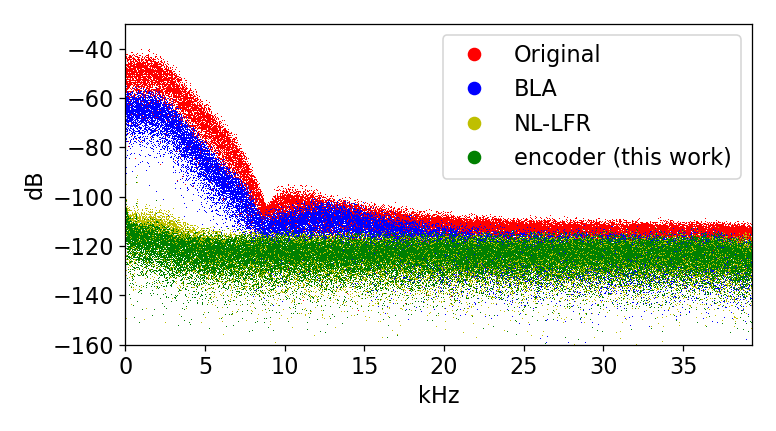}
}
}
\end{figure}

\begin{figure}
\floatbox[{\capbeside\thisfloatsetup{capbesideposition={right,top},capbesidewidth=7.5cm}}]{figure}[\FBwidth]
{\caption{The lowest NRMS simulation error on the validation set during training for the Wiener--Hammerstein benchmark with references of the simulation error approach (Equation \ref{eq:OE}) and using the encoder method without batch optimization.}\label{fig:training}}
{\includegraphics[width=7.5cm]{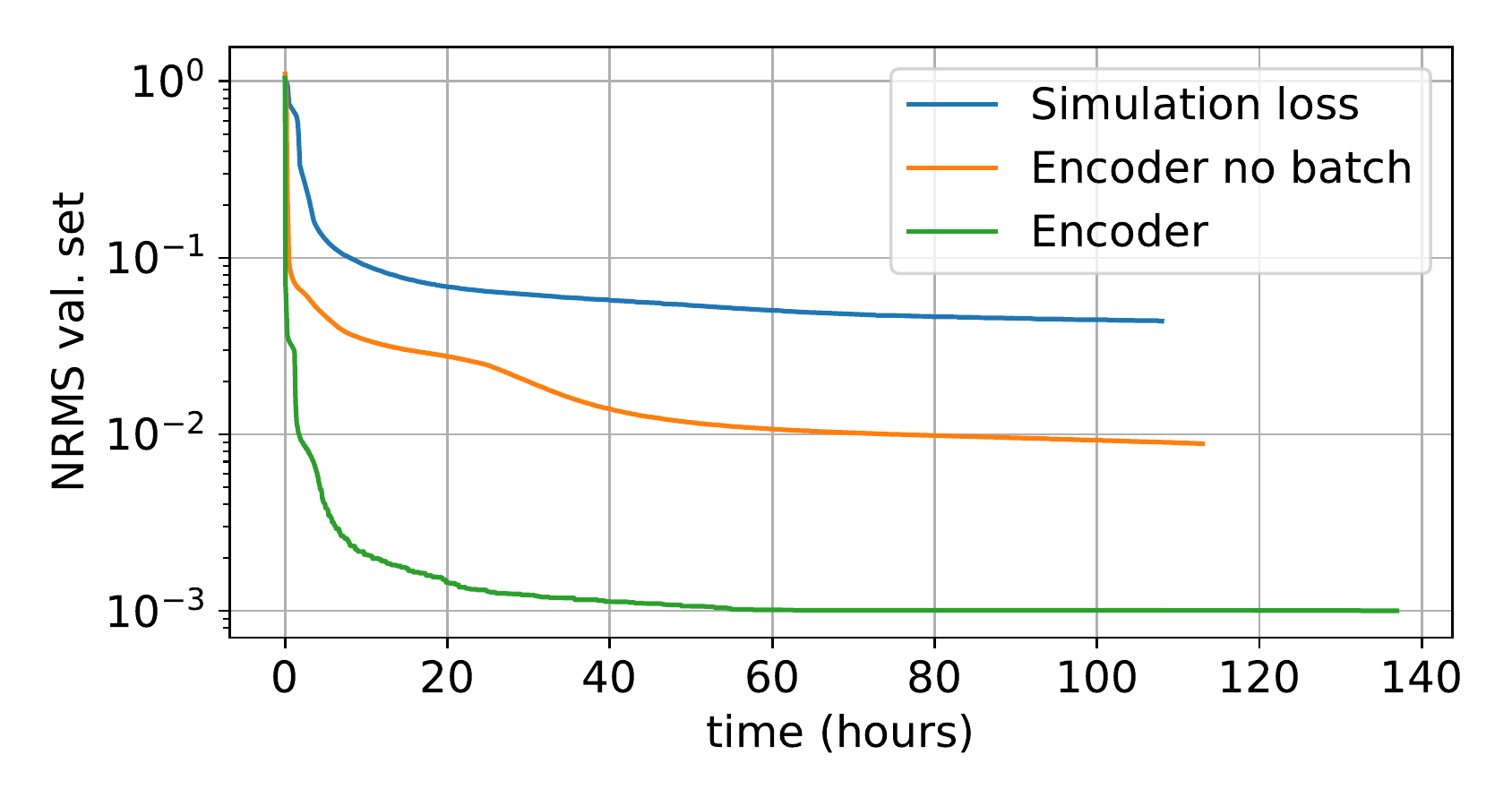}}
\end{figure}

The NRMS error during optimization is shown in Figure \ref{fig:training}. It can be observed that the introduction of encoder-based batch optimalization does not only improve the training time, but also significantly improves the model quality. Though, even with the improvements in training speed introduced by utilizing multiple shooting, the encoder and batch optimization, the training still takes considerable time. The optimization took $4\cdot 10^4$ epochs and $4 \cdot 10^6$ batch updates. However, a high-quality model is already obtained, before the optimization method fully converges, with only a tenth of the time budget. This can possibly be further improved in the future by using more appropriate optimization algorithms such as the recently proposed quasi-Newton methods~\citep{quasi-Newton}.

\begin{figure}
\floatbox[{\capbeside\thisfloatsetup{capbesideposition={right,top},capbesidewidth=7.5cm}}]{figure}[\FBwidth]
{\caption{The $n$-step NRMS as in Equation \eqref{eq:n-step-NRMS} on the test set of the Wiener--Hammerstein benchmark.}\label{fig:n-step}}
{\includegraphics[width=7.5cm]{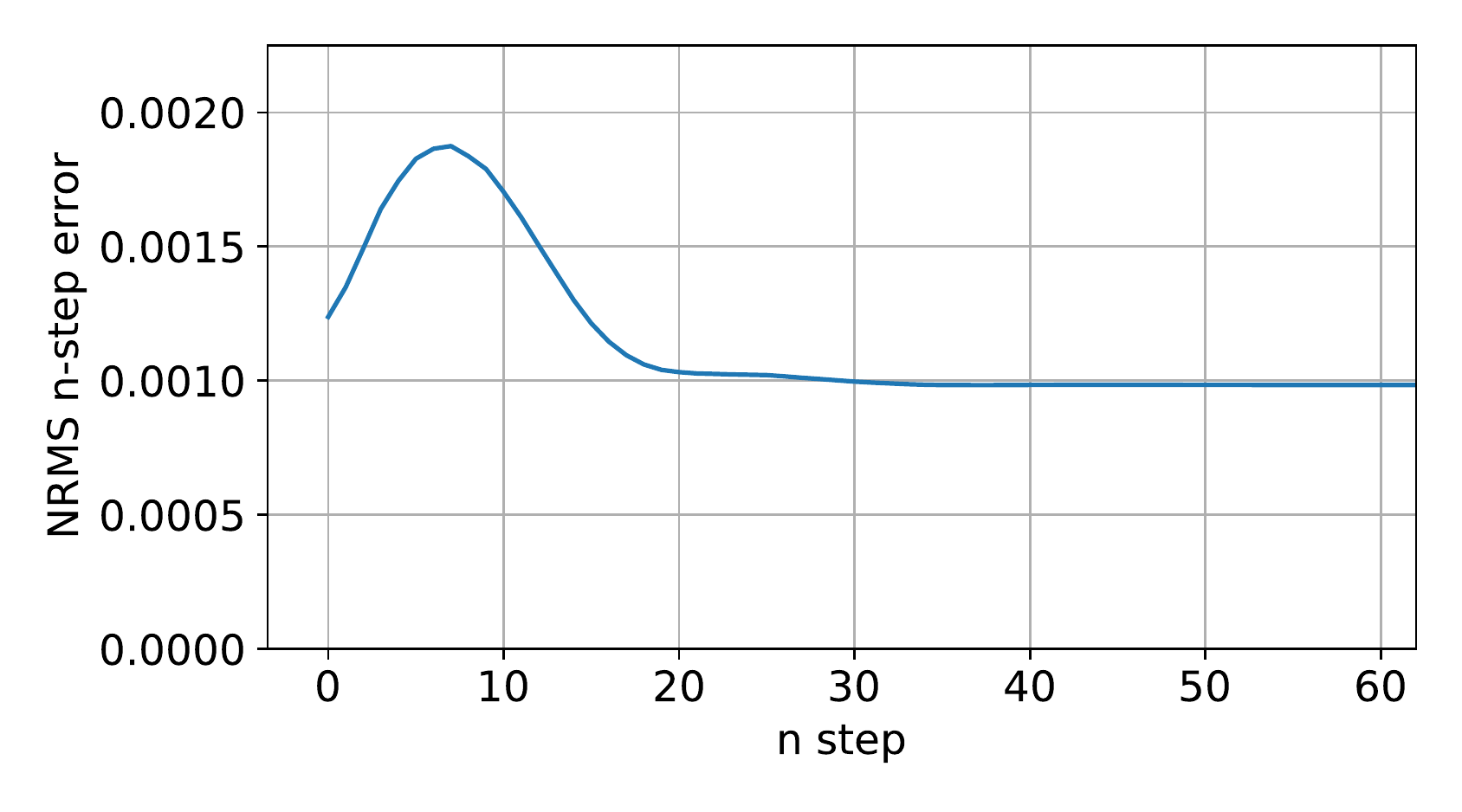}}

\end{figure}

The $n$-step NRMS error is introduced to get some insight into how well the encoder can estimate the initial state and to validate the choice of hyper-parameters $k_0=0$ and $T=80$. The $n$-step NRMS error is introduced as the normalized error you expect so find after taking $n$ steps as in:
\begin{align}
\label{eq:n-step-NRMS}
    \text{NRMS}_n = \frac{\sqrt{1/M \sum_{i=1}^{M} (\hat{y}_{t_i \xrightarrow{} t_i + n} - y_{t_i + n})^2} }{\sigma_y}.
\end{align}
This quantity is shown in Figure \ref{fig:n-step}. The average is taken over all the possible starting $t_i$ of the test set. In the figure, one can see that the encoder does not provide a perfect estimate of the initial state as indicated by a bump that is seen for $n<30$ with the peak being at $n=7$. Nevertheless, this does validate the choice of $T=80$ and $k_0=0$ as the transient does not dominate the loss function. %

\subsection{The Silverbox benchmark}

The Silverbox system benchmark~\citep{benchmark} is an electronic implementation of a mass-spring-damper system with a nonlinear spring, i.e. a forced Duffing oscillator. The Silverbox system can be modeled by a second-order nonlinear state-space model. The data set consists of \textit{test} section of 40000 samples consisting of a filtered Gaussian excitation with slowly increasing amplitude whereas the \textit{train} section of 87000 samples consists of the same filtered Gaussian excitation with constant amplitude where the last 30000 samples are also used to monitor the performance. The remaining 21000 samples composes the validation set. Note that the constant amplitude of the train and validation sections is smaller than the highest amplitude present in the test section requiring good extrapolation properties during testing.

The state-space encoder setup for the Silverbox benchmark is similar to the setup for the Wiener--Hammerstein benchmark. A 2 hidden layer neural network with 64 nodes per layer, $\tanh$ activations, and a linear bypass for $e_\theta$, $f_\theta$ and $h_\theta$ is used. The encoder hyperparameters are set as: $k_0=0$, $T=100$, $n_a = n_b = 50$, a batch size of 256, and using the Adam optimizer with the learning rate $\alpha = 10^{-3}$. The multiple shooting starting point can again be any possible starting point within the range of the training set.

\begin{table}[]
\centering
\caption{Performance of the state-space encoder method on the Silverbox benchmark compared with literature. The numbers in parentheses indicate test set evaluation excluding extrapolation.}
\vspace{-5 mm}
{\small
\begin{tabular}{l|l|l}
\hline
\multicolumn{1}{c|}{\begin{tabular}[c]{@{}c@{}}Identification   \\ Method\end{tabular}} & \multicolumn{1}{c|}{\begin{tabular}[c]{@{}c@{}}Val. RMS\\  simulation \\ (mV)\end{tabular}} & \multicolumn{1}{c}{\begin{tabular}[c]{@{}c@{}}Test RMS \\ simulation\\  (mV)\end{tabular}} \\ \hline
PNLSS \citep{app:PNLSS}                                                                                  &  --                                                                                          & 0.26                                                                                       \\
Sigmoidal network models \citep{app:sigmoidal}                                                               &  --                                                                                          & 0.3                                                                                        \\
LS-SVM \citep{app:LS-SVM}                                                                                  & 0.23                                                                                       & 0.32                                                                                       \\
Poly-LFR  \citep{app:poly-LFR}                                                                              &     --                                                                                       & 0.35                                                                                       \\
Genetic Programming \citep{app:GP-Dhruv}                                                                                   & 0.09                                                                                       & 0.36                                                                                       \\
Direct Identification \citep{app:direct}                                   & 1.4                                                                                        & 0.96                                                                                       \\
Local linear models \citep{app:local-lin}                    & 1.1                                                                                        & 1.3                                                                                        \\
\textbf{State-space Encoder with $n_x=4$ (this work)}                                                                   & \textbf{0.36}                                                                              & \textbf{1.4 (0.32)}                                                                        \\
\textbf{State-space Encoder with $n_x=2$ (this work)}                                                                   & \textbf{1.0}                                                                               & \textbf{2.4 (0.83)}                                                                        \\
Best Linear Approximation  \citep{app:BLA-silver}                                                             & 6.9                                                                                        & 13.5                                                                                      
\end{tabular}
}
\label{tab:silverbox}
\end{table}

\begin{figure}
\floatbox[{\capbeside\thisfloatsetup{capbesideposition={right,top},capbesidewidth=6.5cm}}]{figure}[\FBwidth]
{\caption{The remaining simulation error obtained using the encoder method ($n_x = 4$) on the test set of the Silverbox benchmark. When the output exceeds the training values (i.e. extrapolation) the simulation error increases significantly.}\label{fig:error-time-silverbox}}
{\includegraphics[width=7.5cm]{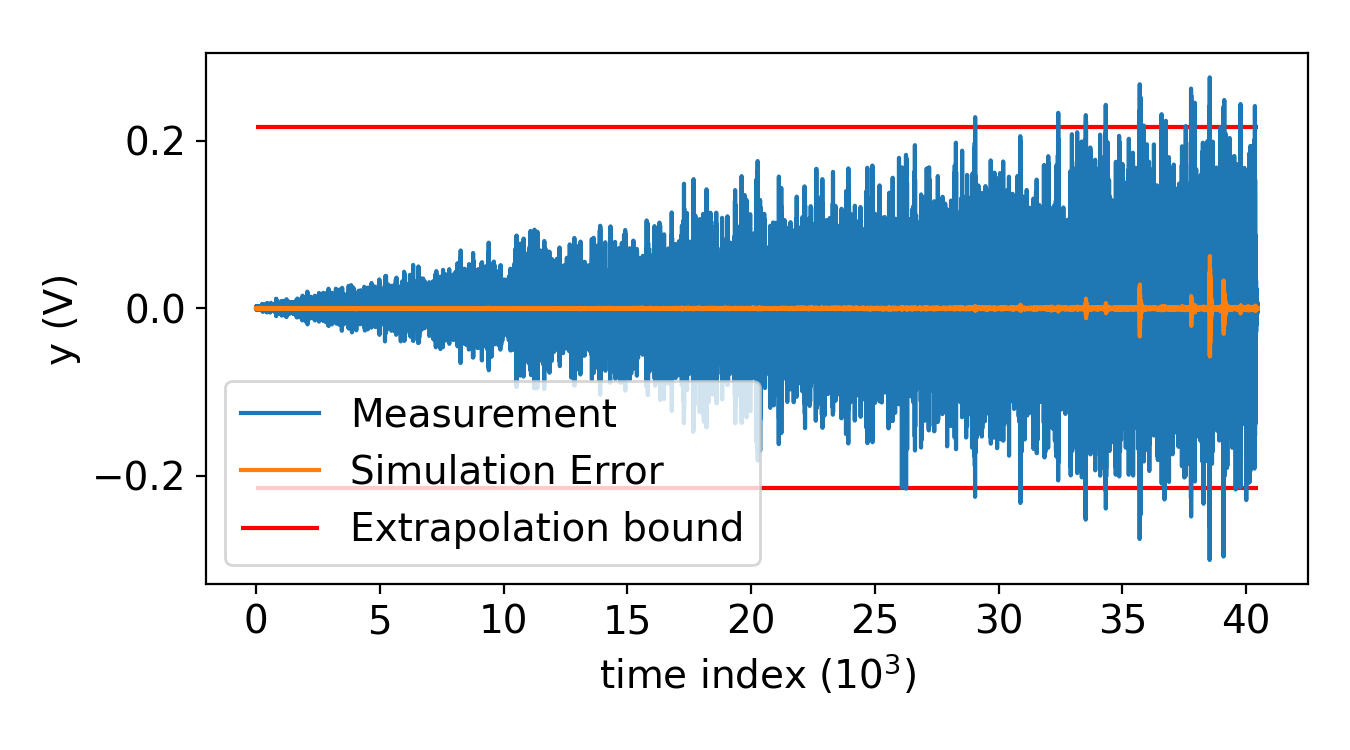}}
\end{figure}

A summary of the results is reported in Table \ref{tab:silverbox}. An observation is that taking the number of internal states equal to the number of real internal states $n_x=2$ performs almost 3 times worse then taking $n_x = 4$. This could be due to the local minima being more pronounced at low state orders. The performance of the state-space encoder method is significantly worse on the test set when comparing to the other methods. However, this can be almost entirely be attributed to the extrapolation errors as can be observed in Figure \ref{fig:error-time-silverbox}. Furthermore, observe that almost all the state-of-the-art models use a polynomial representation of the nonlinearity, which matches with the true system structure. However, the method presented used a neural network to model the nonlinear function which introduced larger extrapolation errors. When the region of extrapolation is excluded from the test set the error drops in the range of state-of-the-art performance.\vspace{-0.33cm}

\section{Discussion}
\label{sec:dis}

This paper presented a novel nonlinear system identification approach realized by combining ideas and methods from machine learning, multiple shooting and subspace identification. The introduction of an encoder function that estimates the internal state based on the historical input and output data, together with multiple computational improvements, such as batch optimization, results in a computationally efficient nonlinear identification method that scales well to large datasets. Without requiring specific parameter initialization approaches (random parameter initialization has been used), the method was able to obtain the best known performance on the Wiener--Hammerstein benchmark. However, currently a known drawback of the encoder method is that the hyper-parameters are system dependent and require manual tuning. A detailed theoretical analysis and further computational improvements of the proposed identification method are the subject of future work. \vspace{-0.5cm}

\setlength{\bibsep}{3.5pt plus 0.3ex}
\bibliography{references}

\end{document}